# Feature Point Extraction for Extra-Affine Image

Tao Wang, Yinghui Wang, Yanxing Liang, Liangyi Huang, Jinlong Yang, Wei Li, Xiaojuan Ning

*Abstract*—The issue concerning the significant decline in the stability of feature extraction for images subjected to large-angle affine transformations, where the angle exceeds 50 degrees, still awaits a satisfactory solution. Even ASIFT, which is built upon SIFT and entails a considerable number of image comparisons simulated by affine transformations, inevitably exhibits the drawbacks of being time-consuming and imposing high demands on memory usage. And the stability of feature extraction drops rapidly under large-view affine transformations. Consequently, we propose a method that represents an improvement over ASIFT. On the premise of improving the precision and maintaining the affine invariance, it currently ranks as the fastest feature extraction method for extra-affine images that we know of at present. Simultaneously, the stability of feature extraction regarding affine transformation images has been approximated to the maximum limits. Both the angle between the shooting direction and the normal direction of the photographed object (absolute tilt angle), and the shooting transformation angle between two images (transition tilt angle) are close to 90 degrees. The central idea of the method lies in obtaining the optimal parameter set by simulating affine transformation with the reference image. And the simulated affine transformation is reproduced by combining it with the Lanczos interpolation based on the optimal parameter set. Subsequently, it is combined with ORB, which exhibits excellent real-time performance for rapid orientation binary description. Moreover, a scale parameter simulation is introduced to further augment the operational efficiency. The experimental results indicate that on the same dataset, the processing speed of the method proposed in this paper is over five times that of ASIFT. Meanwhile, the accuracy of image matching is enhanced by 15% compared with that of ASIFT, and the memory usage is approximately 56% of that of ASIFT. When compared with the latest improved method based on ASIFT, namely Fast-AASIFT, the average time consumption of the method presented in this paper is only 61.4% of that of Fast-AASIFT, the average accuracy is approximately 10% higher, and under the absolute tilt angles ranging from 0 degrees to 80 degrees and the transition tilt angles from 0 to 90 degrees of the viewing angle, the image feature extraction maintains the best performance compared with the current methods.

This work was supported by the National Key Research and Development Program (No. 2023YFC3805901), in part of the National Natural Science Foundation of China (No. 62172190), in part of the "Double Creation" Plan of Jiangsu Province (Certificate: JSSCRC2021532), and in part of the "Taihu Talent-Innovative Leading Talent" Plan of Wuxi City (Certificate Date:202110). (*Corresponding author: Yinghui Wang*)

Tao Wang, Yinghui Wang, Yanxing Liang, Jinlong Yang, Wei Li are with School of Artificial Intelligence and Computer Science, Jiangnan University, Wuxi 214122, China (email: wangyh@jiangnan.edu.cn).
Liangyi Huang is with School of Computing and Augmented Intelligence, Arizona State University, 1151 S Forest Ave Tempe, AZ 8528, USA (email: lhuan139@asu.edu).
Xiaojuan Ning is with Department of Computer Science & Engineering, Xi'an University of Technology, Xi'an 710048, China (email: ningxiaojuan@xaut.edu.cn).

## I. INTRODUCTION

AS a fundamental and critical task in image matching and recognition [9,10,31], feature points extraction is widely utilized in numerous crucial fields, such as remote sensing image processing [1-5,32], object detection [6], and three-dimensional reconstruction [7,8]. Nonetheless, the representation of feature points is intricately intertwined with the information of their surrounding neighborhood regions. Consequently, when an image is subjected to a more pronounced affine transformation, it will instigate significant alterations in the grayscale as well as other pertinent information of both the feature points and their adjacent areas.

Morel and Yu introduced the concepts of absolute tilt and transition tilt [17] to assess the degree of variation in the affine angle of an image. The absolute tilt denotes the slope change that transpires when the camera's viewing angle transitions from a frontal perspective to an oblique one, whereas the transition slope represents the slope between two images with different affine deformations. Additionally, even a marginal difference in absolute tilt can give rise to a substantial transition tilt. Whether the absolute tilt or the transition tilt is large, it implies a considerable amplitude of change in the camera's viewing angle during image capture, which directly impinges on image feature extraction and matching verification. Even the top affine feature extraction methods can't ensure stable extraction of image features under large angles. This is especially salient for large-view images. Once the transition tilt angle exceeds 50 degrees, the stability of feature extraction will undergo a rapid decline, and when this angle surpasses 65 degrees, current image feature extraction methods can scarcely extract stable features. Based on this, researchers have generated a copious number of improved outcomes predicated on classical method [10-12,17,30]. Among them, ASIFT (Affine-SIFT) [17] represents a landmark approach for addressing the issue of extracting feature points from affine transformation images. This method has been demonstrated to be completely affine invariant both theoretically and experimentally. However, the ASIFT algorithm is characterized by a high computational complexity and a protracted computation time [10,17], rendering it arduous to adapt to the application of large-view images. This has become a consensus within the academic community [18,21,34]. Consequently, the pursuit and exploration of methods that can not only meet the high-precision requirements for extracting feature points from images under complete affine transformation, especially in the context of large-view affine transformation images, but also possess real-time performance have been plaguing researchers for years

[22,37,40,41,42]. Simultaneously, it remains a conundrum in the field of computer algorithms where precision and time cost cannot be simultaneously attained.With the aim of addressing these issues, we put forward an image feature point extraction method that focuses on enhancing the accuracy of large-view affine transformation while maintaining speed. The method tackles problems, such as information loss caused by the sudden change in the grayscale of the overlapping areas between images, due to large affine transformations by simulating different viewpoints. Meanwhile, the dual-image simulation strategy of ASIFT is refined into a single-image simulation, and combined with the fast and efficient oriented binary description strategy of ORB (Oriented FAST and Rotated BRIEF) [23] to obtain the optimal parameter set (longitude, latitude and scale) for feature extraction, further augmenting the overall efficiency of the method.

The innovative facets of this paper can be summarized as follows:

(1) A method of single-image affine simulation spatial transformation within the context of the cone viewpoint is proposed. This method undertakes the simulation of affine transformations for images within the three-dimensional cone space. It is not only compatible with traditional image transformations such as translation and rotation in spherical space simulations but also provides support for the scale parameter, thereby significantly enhancing the accuracy of extracting image point features under scaling transformations.

(2) A high-precision matching method for large affine simulation transformation images predicated on the optimal parameter set is devised. This method initially acquires the optimal parameter set that most closely approximates the actual transformation relationship between the two images. Subsequently, it conducts feature extraction on the images based on this optimal parameter set. This approach not only sustains a high level of precision in feature matching but also considerably reduces the running time.

(3) A method for estimating the reference image and the target image with respect to three-dimensional reconstruction is proposed. This method enables the determination of the camera's position and the distance relationship between the centers of the images based on the scaling relationship between the two images. Consequently, it effectively estimates the reference image and the target image. It overcomes the limitation of existing methods that rely solely on known reference image data sets, thereby enhancing the robustness of multi-view three-dimensional reconstruction based on image feature matching.

## II. Related Work

In the domain of addressing affine transformations, the extant methodologies for image feature extraction can be primarily categorized into two principal classes: partial affine transformation and complete affine transformation. The first class encompasses methods such as Harris/Hessian-Affine [11] and MSER (Maximally Stable Extremal Regions) [12, 13]. Within this category, certain algorithms substitute the feature detection step within the SIFT (Scale Invariant Feature Transform) [14] algorithm with an affine invariant feature extraction operator that exhibits enhanced performance [15]. Meanwhile, other algorithms utilize the second-order gradient moment, which represents the local shape, to normalize the SIFT features, thereby alleviating the impact of affine transformations [16]. To a certain extent, this mitigates the problem where SIFT fails to extract stable features across different scale spaces due to substantial image transformation. However, given that the initial scale selection and feature positioning in the existing affine invariant feature extraction algorithms do not rely on a fully affine invariant approach [11], this class of methods is incapable of attaining true complete affine invariance and demonstrates insufficient adaptability to more pronounced affine deformations.

The second class of methods, exemplified by the ASIFT algorithm that supports complete affine transformations, offers a novel perspective for resolving the affine transformation dilemma. The ASIFT algorithm draws inspiration from the exhaustive strategy of simulating the scale space employed by SIFT. It achieves complete affine invariance by modifying the longitude and latitude of the direction of the camera's principal optical axis, thereby simulating images from diverse viewing angles within the constraint of the hemisphere. Subsequently, feature extraction and matching are conducted on the acquired images using a scale-invariant method. Both empirical and theoretical evidence has substantiated its complete affine invariance. Nevertheless, the ASIFT algorithm is beset with issues such as a substantial computational burden and scale invariance within the hemisphere constraint.

Simultaneously, in an attempt to address the issue of the extensive computational time required by the ASIFT algorithm, on the basis of the ASIFT algorithm, the AORB (Affine-ORB) algorithm [18], the ASURF (Affine-SURF) algorithm [19, 20], and the AFREAK (Affine-FREAK) algorithm [21, 22] respectively employ the ORB, the SURF (Speeded Up Robust Features) [24, 25], and the FREAK (Fast Retina Keypoint) [26] algorithms to replace the SIFT algorithm, thereby effectively enhancing the operational speed of the algorithm. However, each of these algorithms is fraught with certain deficiencies. For instance, the AORB method, which is based on the ORB algorithm renowned for its excellent real-time performance, exhibits a significant disparity in accuracy when compared to the SIFT algorithm. This shortcoming results in the feature extraction and matching efficacy of the AORB algorithm under affine transformation being far inferior to that of the ASIFT. The ASURF method utilizes the SURF algorithm, which, by reducing the dimension of the feature description operator to 64 dimensions on the basis of SIFT, enables ASURF to operate at a faster pace and with greater efficiency than ASIFT. However, this also leads to a slight decline in the stability of ASURF under affine changes, and there remains a certain disparity in speed when compared to the AORB, rendering it incapable of fulfilling the application requirements of certain fields with real-time demands. The AFREAK algorithm operates at a relatively rapid pace and exhibits relatively good accuracy, yet the number of effective feature points it generates is relatively small, and the number of effective feature points it extracts under larger-angle affine transformations is far less than that of the ASIFT. The improved ASIFT method put forward by Gao et al. [34], which incorporates the known POS (Position and Orientation System) information of UAVs (unmanned aerial vehicles), harnesses the POS data to acquire simulated images and curtail sampling parameters. Subsequently, the SIFT algorithm in conjunction with BRISK (Binary Robust Invariant Scalable Keypoints) descriptors [36] is utilized to extract image features. In certain scenarios, the time consumption of this method is approximately half that of the ASIFT method, while its accuracy is around 10% higher than that of ASIFT. Nevertheless, this approach suffers from significant limitations. In the majority of practical application scenarios, it poses a formidable challenge to obtain the POS information beforehand. Qi [35] proposed a method whereby the feature point pairs within the images are first extracted via SIFT feature extraction, thereby deriving the transformation matrix between the images. Subsequently, a single perspective transformation is carried out on the affine images, followed by another round of SIFT feature extraction. This method, hereinafter abbreviated as Qi-SIFT for the sake of simplicity, is capable of obtaining a greater number of feature points compared to the SIFT method and exhibits a much higher efficiency than ASIFT, with the time consumption being approximately one-ninth that of ASIFT. However, this method fails to resolve the issue of determining which image should be selected for transformation in practical applications. Moreover, in scenarios where the SIFT algorithm becomes inoperable, Qi-SIFT will distort the images due to incorrect homography matrices, and the results of this method will be inferior to those of SIFT under large-view affine transformations.

Overall, the ASIFT method demonstrates the best affine invariance and matching accuracy, yet it lags behind in terms of speed performance. The underlying reason is that ASIFT calculates the affine deformation matrix corresponding to pose parameters at different sampling points, generates reference and target image simulation groups based on the affine matrix, and then utilizes an image matching algorithm to match the affine-transformed images one by one to obtain the final result. The ASIFT's image simulation strategy is only capable of resolving the image matching problem caused by the change of the camera's principal optical axis (i.e., only handle affine transformation related to longitude and latitude changes within the semi-circular spherical surface), lacking other degrees of freedom. In contrast, the SIFT algorithm can meet the relevant requirements. When the ASIFT method is combined with the SIFT method, it attains a higher accuracy by simultaneously performing simulation affine transformation on both reference and target images to obtain two groups of simulated images. However, due to the processes required by the SIFT algorithm, such as generating Gaussian pyramids and difference pyramids and establishing 128-dimensional feature vectors, the ASIFT algorithm consumes more memory and has a high time complexity.

It can be discerned that the ASIFT method, which is predicated on the SIFT approach, exhibits complete affine invariance and attains a relatively high level of accuracy in the extraction and matching of image features under affine transformation. However, it is marred by a low temporal efficiency. In contrast, the AORB method, which is based on the ORB algorithm, is characterized by its rapid processing speed, albeit with a computational accuracy that leaves something to be desired. If the dual-image simulation affine transformation strategy of ASIFT were to be converted into a single-image simulation, with the feature extraction still being carried out on the basis of ORB, and the scale transformation parameter were to be incorporated into the set of simulated transformation parameters, it would serve to remedy the shortcoming of the ORB algorithm's relatively poor scale invariance. This, indeed, constitutes the central tenet of the method expounded in this paper.

## III. Method

The framework of our method is depicted in Fig. 1, primarily comprising three major constituents: the selection of reference images, the acquisition of the optimal set of simulation parameters, and the precise matching process predicated on the said parameter set.

The primary challenge in this paper resides in ascertaining which image ought to undergo affine simulation transformation to attain the most favorable outcomes. To this end, SIFT feature extraction and matching procedures are conducted on the two original images. A subset of feature points is selected from Image 1 and paired to constitute line segments. For the corresponding matching points in Image 2 that align with those in Image 1, analogous line segments are also interconnected thereon and deemed as matching line segments. In this manner, the scaling coefficient is derived through the weighted summation of the length differences between pairs of matching line segments. The sign and magnitude of the scaling coefficient serve to reflect the distance relationship between the camera positions corresponding to the two input images and the image centers within the three-dimensional space. Based on this, the reference image and the target image can be demarcated.

Secondly, rapid simulation affine transformation is executed on the reference image, founded on diverse parameter sets (encompassing longitude parameters, latitude parameters, and scale parameters) in conjunction with the nearest-neighbor interpolation method. This process yields image groups corresponding to distinct sampling point positions. ORB feature extraction and matching operations are then carried out using each image within the image group and the target image, with the parameters corresponding to the maximal number of obtained matching point pairs being designated as the optimal sampling parameter set.

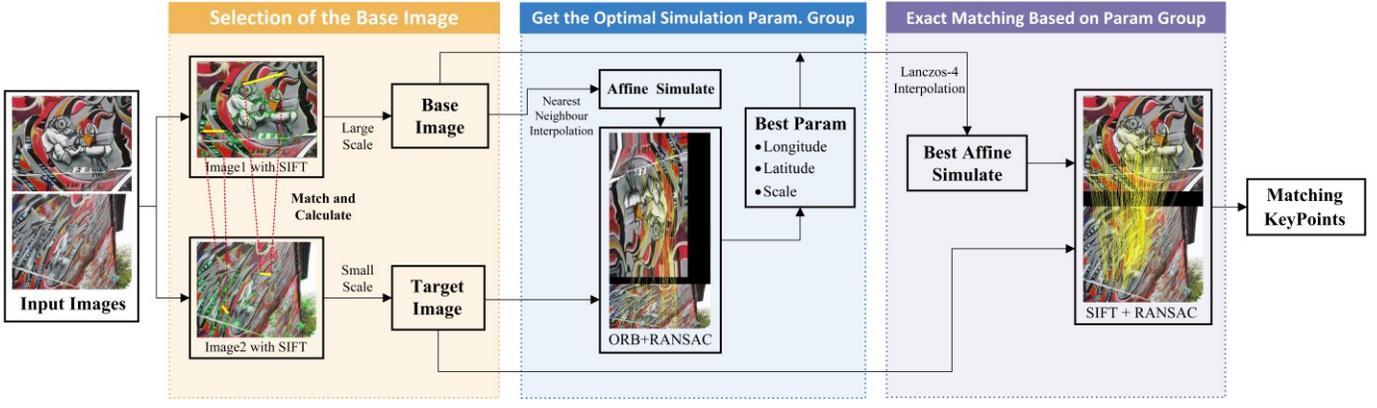

**Fig. 1.** Schematic diagram of the technical framework for the method.

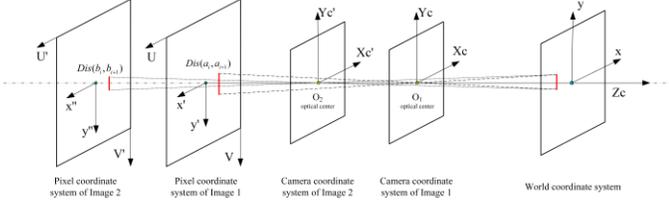

**Fig. 2.** Schematic diagram of the correspondence relationship of camera imaging positions.

Finally, predicated on the optimal sampling parameter set, affine simulation transformation is effected on the reference image by means of Lanczos interpolation [27]. Subsequently, the SIFT algorithm is enlisted to undertake feature extraction and matching maneuvers between the transformed reference image and the target image, with the objective of further refining the matching results. Through this sequence of operations, not only are the feature points on the images successfully retrieved, but also the stability of the extracted feature points can be corroborated through the matching of feature point pairs between the reference image and the target image.

*A. Selection of The Base Image*

Denote the set of feature points on Image 1 as $A=\{a_1,a_2,a_3,\ldots,a_n\}$, and the set of feature points on Image 2 corresponding to those in $A$ as $B=\{b_1,b_2,b_3,\ldots,b_n\}$. Theoretically, let $<a_i, a_{i+1}>$ ($1 \le i \le n-1$) represent a line segment on Image 1, which also corresponds to a certain line segment in the three-dimensional space, and $Dis(a_i, a_{i+1})$ represents the pixel length of this line segment in Image 1. Meanwhile, let $<b_i, b_{i+1}>$ denote a line segment on Image 2, with $Dis(b_i, b_{i+1})$ representing the pixel length of this line segment in Image 2, as illustrated in Fig. 2.

The formula for calculating the scaling coefficient f from Image 1 to Image 2 is shown in (1).

$$f = \sum_{i=0}^{4i+1<n}\left(\left(Dis(a_i,a_{i+1})-Dis(b_i,b_{i+1})\right)\times\left(1-\frac{Des(i)+Des(i+1)}{1000}\right)\right) \quad (1)$$

In (1), $Des(i)$ represents the descriptor distance when performing SIFT feature matching between feature point $a_i$ in Image 1 and its matching point $b_i$ in Image 2, and $Des(i+1)$ represents the descriptor distance between point $a_{i+1}$ in Image 1 and point $b_{i+1}$ in Image 2.

As the descriptor distance represents the credibility of the matching relationship between this pair of feature points, the smaller the descriptor distance is, the more reliable the matching relationship of the feature point pair will be. In (1), the sum of the descriptor distances $Des(i)+Des(i+1)$ is used as the weight of the length difference $Dis(a_i, a_{i+1})-Dis(b_i, b_{i+1})$ between line segment $<a_i, a_{i+1}>$ and line segment $<b_i, b_{i+1}>$ on Image 2. Specifically, the smaller the sum of descriptor distances associated with a particular point, the larger the corresponding weight. This implies that line segment pairs with more dependable matching relationships at both endpoints exert a more significant influence on the scaling coefficient, thereby yielding more precise results.

When $f > 0$, it manifests that for the identical line segment within the three-dimensional space, its mapped length in Image 1 surpasses that in Image 2. This indicates that the optical center of the camera corresponding to Image 1 is in closer proximity to the actual target. In comparison with Image 2, Image 1 encompasses more detailed information. Hence, Image 1 is more apt to be employed as the reference image for conducting simulated affine transformations. Upon undergoing simulated affine transformations, a greater number of key features will be preserved, which is more conducive to the final experimental results. In practical applications, with the aim of augmenting the robustness of the method, the scaling coefficient $f_1$ can be computed from Image 1 to Image 2, and subsequently, the scaling coefficient $f_2$ can be calculated from Image 2 to Image 1. When $f_1 > f_2$, it can be presumed that the distance between the optical center of the camera corresponding to Image 1 and the center of the image is smaller. Based on this premise, Image 1 is designated as the reference image and Image 2 as the target image, and vice versa.

*B. Construction of The Optimal Simulation Parameter Set*

Through the first-order Taylor formula, the smooth deformation of any plane can be approximately solved and processed for each point [17]. The surface deformation of the target object caused by the change in camera position can be modeled by local affine transformation through affine transformation. The local transformation of the camera is equivalent to multiple cameras at an infinite distance, and these infinitely distant cameras will produce affine deformation. Fig. 3 shows the corresponding camera position movement process of affine decomposition. When $s_0 = 1$ and $\theta_0 = \phi_0 = \psi_0 = 0$, the camera is positioned right above the image. As the parameter groups $s, \theta, \phi, \psi$ change, the camera position will also undergo corresponding rotational or translational movements. Based on the above principles, we can alter the camera position by changing the parameter groups, which is also the mechanism by which the simulated affine transformation of the images is carried out in this paper.

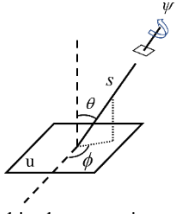

**Fig. 3.** Geometric relationship between image affine transformation and camera position. The image $u$ is a physical object on a plane. The parallelogram in the upper right corner represents the camera observing $u$. $\psi$ represents the spin of the camera. $s$ is the distance from the optical center of the camera to the center of the image. Both the longitude angle $\theta$ and the latitude angle $\phi$ are the angles between the viewpoint and the optical axis.

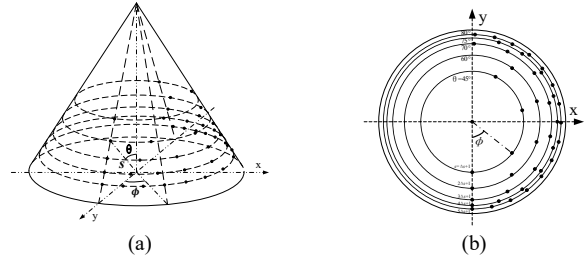

**Fig. 4.** Illustration of the distribution of sampling points. (a) shows the distribution of sampling points in the spatial coordinate system; (b) presents the distribution of the projections of sampling points on the xoy plane.

Therefore, the transformation of the camera position can be converted into an affine transformation matrix with a positive determinant and has a unique decomposition, as shown in (2).

$$A = T R_1(\psi) S T_\tau R_2(\phi) \quad (2)$$

In which,

$$T = \begin{pmatrix} 1 & 0 & t_x \\ 0 & 1 & t_y \\ 0 & 0 & 1 \end{pmatrix}, R_1(\psi) = \begin{pmatrix} \cos\psi & -\sin\psi & 0 \\ \sin\psi & \cos\psi & 0 \\ 0 & 0 & 1 \end{pmatrix}, S = \begin{pmatrix} s & 0 & 0 \\ 0 & s & 0 \\ 0 & 0 & 1 \end{pmatrix},$$

$$T_\tau = \begin{pmatrix} t & 0 & 0 \\ 0 & 1 & 0 \\ 0 & 0 & 1 \end{pmatrix} = \begin{pmatrix} 1/\cos\theta & 0 & 0 \\ 0 & 1 & 0 \\ 0 & 0 & 1 \end{pmatrix}, R_2(\phi) = \begin{pmatrix} \cos\phi & -\sin\phi & 0 \\ \sin\phi & \cos\phi & 0 \\ 0 & 0 & 1 \end{pmatrix}$$

$T$ represents the translation transformation matrix, where $t_x$ and $t_y$ denote the translation variables. $R_1(\psi)$ is the camera spin transformation matrix, and $\psi$ represents the spin angle of the camera. $S$ is the scaling transformation matrix, and $s$ is the distance from the optical center of the camera to the center of the image. $T_\tau$ is the tilt matrix, and $t$ is the tilt degree. The latitude angle $\theta = arccos 1/t$. $R_2(\phi)$ is the rotation matrix, and $\phi$ is the longitude angle. Both the latitude angle $\theta$ and the longitude angle $\phi$ are the angles between the viewpoint and the optical axis.

In order to ensure that the simulated images maintain similarity under different parameter groups, during the coarse matching process, sampling is carried out in accordance with specific rules in this paper. The sampling value of the tilt degree corresponding to $\theta$ is denoted as $t = 1, a, a^2, \ldots, a^n$, which required that $a > 1$, and $a = \sqrt{2}$, $n = 5$ in this particular paper. The longitude angle $\phi$ is assigned values from an arithmetic sequence of the latitude angle, namely 0, $b/t$, $2b/t$, …, $kb/t$. Meanwhile $b$ is set to 72°, and the value of $k$ is taken as the largest natural number that satisfies the condition $kb/t < 180°$. Regarding the scale $s$, its value is such that $s = 1+\Delta s, 1+2\Delta s, \ldots, 1+n\Delta s$. Moreover, the coefficient of $\Delta s$ is made to be consistent with the exponent of the tilt degree $t$, that is, $n = 5$. The distribution of sampling points is illustrated in Fig. 4. Evidently, the sampling mode based on Fig. 4 is distinct from the ASIFT sampling mode which is based on the hemisphere.

During the procedures of image simulated affine transformation and image feature extraction, this paper utilizes the nearest neighbor interpolation method and the ORB feature extraction algorithm. Moreover, it combines these with the Brute-Force Matcher based on the Hamming distance lookup table, with the aim of maximizing the operational speed within the coarse matching phase. To address the drawback that the ORB algorithm demonstrates suboptimal performance under substantial scale variations, a scale variable is incorporated into the sampling parameter group for the simulated affine transformation in this study. When the magnitude of the simulated affine transformation is relatively large, an appropriate scale amplification transformation is implemented, which can effectively compensate for the loss of image details that occurs during extensive affine transformations. Subsequently, the longitude angle, latitude angle, and scale corresponding to the simulated image possessing the maximum number of matching points are identified and recorded. At this juncture, these longitude angle, latitude angle, and scale parameters can form the optimal sampling parameter group, which will be employed for the subsequent precise matching.

### C. Precise Matching Based on Parameter Group

The SIFT algorithm, in essence, is to detect feature points and calculate their directions within different scale spaces, which endows it with stable matching accuracy and favorable affine invariance. In the method proposed in this paper, the SIFT algorithm is an ideal approach for accurate matching.

After the optimal set of sampling parameters is obtained, the reference image is used to reproduce the simulated transformation with this parameter set, and SIFT feature extraction and matching are carried out with the target image.

In order to preserve details and minimize image distortion during the process of size change, Lanczos interpolation [27] is adopted during the simulated transformation, and this method is implemented based on (3).

$$f(x,y) = \sum_{i=-n+1}^{a} \sum_{j=-n+1}^{a} f(\lfloor x \rfloor + i, \lfloor y \rfloor + j) \cdot L(i - x + \lfloor x \rfloor; a) \cdot L(j - y + \lfloor y \rfloor; a) \quad (3)$$

Among them, $\lfloor x \rfloor$ denotes the floor value of $x$, and $a$ is the width of the kernel function, which determines the window size of the interpolation function algorithm. The kernel function is defined as presented in (4).

$$L(x;a) = \begin{cases} \sin c(x) \sin c(x/a) & -a < x < a \\ 0 & \text{otherwise} \end{cases} \quad (4)$$

In which $\sin c(x) = \sin(\pi x)/(\pi x)$.

Lanczos interpolation exhibits remarkable characteristics in terms of detail preservation. Compared with linear interpolation and bicubic interpolation, it can attain an effect approximating that of bicubic spline interpolation at a speed comparable to that of bilinear interpolation. Moreover, it can preserve small-sized structures to the greatest extent possible, which holds substantial significance for subsequent matching procedures. Upon the completion of SIFT matching, the RANSAC (Random Sample Consensus) [28] algorithm is employed to compute the homography matrix and to screen and eliminate the mismatched points, thereby further optimizing the final outcome.

## IV. EXPERIMENTAL RESULTS AND ANALYSIS

This section primarily demonstrates data sets, the standards of evaluation, experimental results, experiments on parameter variations, and the comparative analysis of methods. The code in this paper is developed based on Visual Studio Code and OpenCV 4.5.5. All experiments are conducted within an environment consisting of an Intel Core i7 2.7 - GHz processor and a 6 - GB - RAM Ubuntu 20.04 system installed in VMware Workstation Pro.

*A. Data Sets*

The Mikolajczyk public data set [29] is a standard data set widely used in the field of computer vision for evaluating the performance of image feature extraction and matching algorithms. The data set consists of a series of images in different scenarios, covering various viewpoints, rotations, and affine transformations. Moreover, it also provides the true homography matrices of the transformations. It is a classic data set in affine transformation experiments and can well test the ability of methods to extract and match image features in different complex scenarios. The MorelYu_2009 data set is a self-made data set by the authors of the ASIFT algorithm. It is the only public data set that contains absolute and transition tilts [17] and can be used as a supplementary part of the Mikolajczyk standard data set.

To verify the effectiveness of the proposed method in this paper, first, on the basis of several image sets from the Mikolajczyk dataset and the MorelYu_2009 dataset, a matching verification experiment concerning the effectiveness of feature point extraction is performed. Second, although standard datasets can provide a relatively objective evaluation environment, such data are often collected and processed under ideal conditions (such as consistent illumination conditions). To be closer to the practical application scenarios and further comprehensively evaluate the performance of the proposed method in this paper under complex environmental scenarios, we also take a set of actual photos for feature extraction. The absolute tilt angle of this group of pictures is expanded from 0° to 80°, which is used to verify the adaptability of the proposed method in this paper for large - view - change images taken in practice.

*B. Standards of Evaluation*

In the experiment, using accuracy and running time as evaluation indicators can fully meet the requirements for the result evaluation of the proposed method in this paper and the comparative evaluation.

1) **Precision rate**

The calculation of the accuracy of feature extraction points between Image 1 (reference image) and Image X (image captured under a varying angle) is shown in (5).

$$Precision = (R\_Num / Pt\_Num) \times 100\% \quad (5)$$

Among them, $Pt\_Num$ is the number of feature points for which a matching relationship is established in Image 1, and $R\_Num$ is the number of feature points in Image X that satisfy the condition as shown in (6).

$$\begin{gathered} (x_1, y_1)^T \cdot H_{1\_X} = (x_2', y_2')^T \\ \sqrt{(x_2' - x_2)^2 + (y_2' - y_2)^2} \leq Th \end{gathered} \quad (6)$$

In (6), $(x_1, y_1)$ is the coordinate of the feature point in Fig. 1, $H_{1\_X}$ represents the homography matrix of the real transformation relationship from Image 1 to Image X, $(x_2', y_2')$ is the coordinate of the real point in image X calculated by the matrix for $(x_1, y_1)$, $(x_2, y_2)$ is the coordinate of the point in image X that matches $(x_1, y_1)$ in Image 1 through our method; $Th$ represents the matching error threshold. In the experiments of this paper, except for special instructions, the default value is $\sqrt{3}$, indicating that when the matching point is within the 3*3 grid range centered on the real point, it is considered as a correctly matched point.

2) **Inlier ratio**

Since some data sets lack the real homography matrix, it is impossible to calculate the real coordinates corresponding to the first image, and thus the accuracy rate cannot be calculated. For the MorelYu_2009 data set, we judge the effectiveness of the method by visually observing the matching result graph, and calculate the proportion of RANSAC inliers through (7) as the accuracy index of various methods under different absolute and transition tilt angles. For the real - shot images, we mainly rely on the matching result graph. At the same time, we use the finally calculated feature points to obtain the corresponding homography

TABLE I

TABLE
EXPERIMENTAL RESULTS OF SOME IMAGE SETS IN THE MIKOLAJCZYK DATA SET

| Image Set | Image Pair | Points Num↑ | Precision↑ | Times(ms)↓ |
|---|---|---|---|---|
| Graffiti (Viewpoint change) | 1-2 | 774 | 94.06% | 7228.41 |
| | 1-3 | 769 | 96.88% | 6891.55 |
| | 1-4 | 519 | 95.76% | 6922.59 |
| | 1-5 | 305 | 91.15% | 6920.03 |
| | 1-6 | 240 | 93.33% | 6933.49 |
| Wall (Viewpoint change) | 1-2 | 5064 | 76.78% | 9479.64 |
| | 1-3 | 3970 | 95.84% | 9159.36 |
| | 1-4 | 3032 | 81.00% | 8741.35 |
| | 1-5 | 1415 | 78.80% | 8676.77 |
| | 1-6 | 707 | 77.09% | 8653.90 |
| Bark (Zoom+rotation) | 1-2 | 593 | 67.45% | 6996.25 |
| | 1-3 | 532 | 39.10% | 7407.60 |
| | 1-4 | 681 | 92.36% | 7248.53 |
| | 1-5 | 430 | 99.77% | 7269.13 |
| | 1-6 | 265 | 90.57% | 7378.23 |
| Boat (Zoom+rotation) | 1-2 | 1967 | 96.59% | 7735.00 |
| | 1-3 | 1661 | 96.45% | 7552.43 |
| | 1-4 | 979 | 91.73% | 7574.24 |
| | 1-5 | 617 | 90.92% | 7235.10 |
| | 1-6 | 396 | 89.14% | 7689.46 |

*"1-X" indicates that the method proposed in this paper is applied to Image 1 and Image X of the corresponding image set.

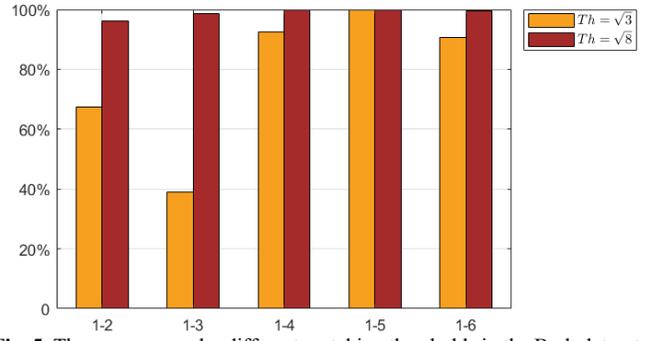

**Fig. 5.** The accuracy under different matching thresholds in the Bark dataset.

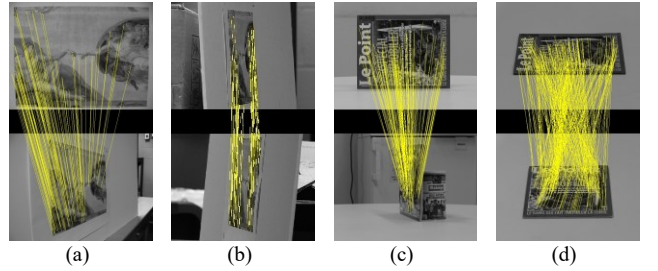

(a)     (b)     (c)     (d)

**Fig. 6.** The matching effect of our method on different image sets of the MorelYu_2009 dataset. (a) Adam painting (absolute tilt 75°L); (b) Adam painting (transition tilts 80°L to 80°R); (c) Magazine (absolute tilt 80°R); (d) Magazine (transition tilt 90°).

matrix and perform perspective transformation on the image, and then analyze the final restored effect graph. Since parameters such as the accuracy rate cannot be accurately calculated, they are only used as a reference for verifying the effect of our method.

$$Ratio_{Inliers} = (In\_Num/Pt\_Num) \times 100\% \quad (7)$$

Among them, *In_Num* is the number of inliers in Image 1 that have been screened by RANSAC.

3) **Running time**

The running time includes the feature extraction time and the feature matching time.

*C. Experimental Result*

During the experiment, the parameter settings for the sampling of our method are as follows: the reciprocal of the inclination $a = \sqrt{2}$, the upper limit *n* of the geometric progression of *a* is equal to 5, and the threshold of the ratio of the descriptor distances between the nearest neighbor and the second nearest neighbor for matching screening is 0.75. In the process of SIFT precise matching, the parameter *n* of the Lanczos interpolation used is 4, that is, the range of nearby pixels used is 8*8. Regarding the number of feature points, the number of feature points extracted on the image during the estimation of the scaling relationship does not exceed 1,000, and the number of feature points extracted in the ORB coarse matching stage is 1,000 for each image. All the above parameters are the optimal ones verified through experiments.

The experimental results of our method on the Graffiti, Wall, Boat and Bark image sets are shown in Table 1.

As is presented in Table 1, the number of feature points and the precision achieved by the method proposed in this paper vary in accordance with the different extents of image transformation. The minimum number of feature points remains steadily above 200, which suffices to support the 3D reconstruction work. The time consumed for calculation generally ranges from 6,000 to 8,000 ms, basically fulfilling the actual requirements. Concerning the accuracy of feature points, our method attains an accuracy of over 90% on most datasets, with some reaching around 80%. Upon observing the data in Table 2, it can be noticed that the precision of two images within the Bark image set is rather low. The reason lies in the fact that the images in this set are primarily composed of soil and plants. Specifically, the proportion of plants in Image 2 and Image 3 is the lowest, while that in Image 4 and Image 5 is relatively high. This circumstance results in poor feature recognizability for Image 2 and Image 3, and consequently, it becomes more challenging to extract features accurate to a single pixel after interpolation when the method our method simulates affine transformation. To further verify the precision of the method on this image set, we adjusted the threshold *Th* in (5) to $\sqrt{8}$. In other words, the correctly matched area was modified to a 5*5 pixels area centered around the real point. The results are illustrated in Fig. 5.

It can be found from Fig. 5 that after the threshold is adjusted to $\sqrt{8}$, the accuracy of the vast majority has been improved to 95% or above. Even for the poorly performing image pair 1-2 and image pair 1-3, the accuracy can also reach a relatively good level, indicating that the vast majority of points are within the 5*5 pixels area centered on the real points, which can basically meet the actual requirements.

Fig. 6 shows the effect of our method on the absolute inclination image set and the transition inclination image set of the MorelYu_2009 dataset. According to the results, whether it is a large absolute tilt or a large transition tilt (the stability

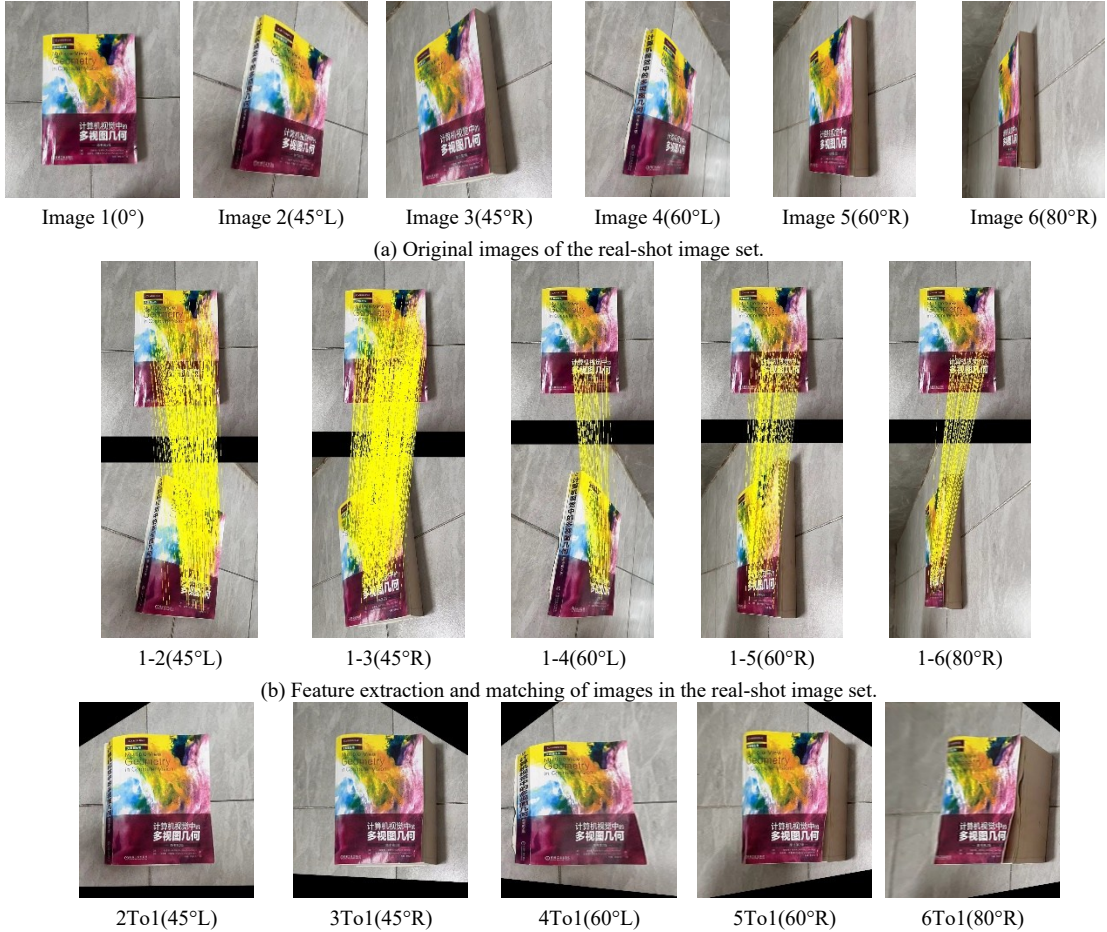

(a) Original images of the real-shot image set.

(b) Feature extraction and matching of images in the real-shot image set.

(c) Effect image of restoring Image 1 (0°) based on the homography matrix calculated from the feature point sets of each image.

**Fig. 7.** The effect of feature extraction and matching of our method on the real-shot image set.

problem of image feature extraction under extremely large angles is also a problem that other existing methods have failed to solve for a long time), our method can achieve high matching accuracy, which also verifies that our method has good stability in extracting feature points.

Fig. 7 illustrates the presentation effect of our method on the real-shot image set. Among them, the transformation angle of Image 6 is a transitional tilt angle of approximately 80°. In practical application scenarios, generally, two tilted views are compared and analyzed, so the transitional tilt is more widely used. Judging from the results of this image set, our method performs well on real-shot images. Within the range where the transitional tilt angle is between 0° and 80°, good experimental results can be achieved. It is able to obtain sufficient feature points and restore the transformation matrices between images through these feature points.

*D. Experiments on Parameter Variation*

This subsection aims to explore how alterations in key parameters affect the precision and speed of our method, thereby facilitating the identification of the parameter values that yield the optimal performance. Primarily, there are two parameters within our method that exert a substantial influence on the results. These are the number of feature points extracted via the ORB method during the acquisition of the optimal simulation parameter group and the value of the variation in the scale parameter. In this study, parameter variation experiments were carried out on the Graffiti image set of the Mikolajczyk standard dataset to examine the impact of these two factors on the precision and speed of our method.

Firstly, an experiment was designed to investigate the influence exerted by different numbers of feature points extracted by ORB during the process of obtaining the optimal simulation parameter group on both the precision and speed of the proposed method. Under the identical variation of the scale parameter, we separately examined the scenarios where the number of feature points extracted each time was set to 500, 800, 1000, 1200, and 1500 respectively. The experimental results concerning precision are presented in Table 2, while those regarding speed are depicted in Fig. 8.

As depicted in Table 2, the number of ORB feature points has a certain impact on the precision. Specifically, in the affine transformations of Images 1- 6, relatively low precision levels are observed when the number of feature points is set at either 500 or 800. The quantity of ORB feature points primarily influences the matching outcomes within the simulated transformations, which, in turn, has implications for the acquisition of the optimal simulation parameter group. Upon entering the subsequent precise matching phase, variations in the input parameter groups lead to differences in

TABLE II

RESULTS OF PARAMETER VARIATION EXPERIMENTS (PRECISION - NUMBER OF ORB FEATURE POINTS)

| Number of feature points | 1-2↑ | 1-3↑ | 1-4↑ | 1-5↑ | 1-6↑ |
|---|---|---|---|---|---|
| 500 | 94.06% | 96.88% | 95.76% | 91.15% | 92.54% |
| 800 | 94.06% | 96.88% | 95.76% | 91.15% | 92.54% |
| **1000** | **94.06%** | **96.88%** | **95.76%** | **91.15%** | 93.33% |
| 1200 | 94.06% | 96.88% | 95.76% | 91.15% | 93.33% |
| 1500 | 94.06% | 96.88% | 95.76% | 91.15% | 93.33% |

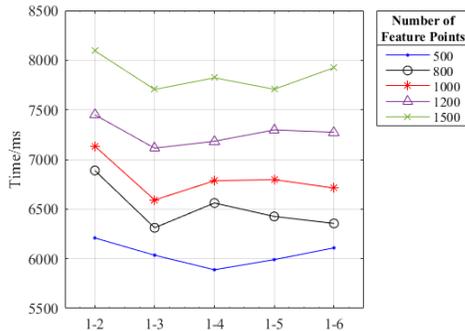

**Fig. 8.** Comparative experiment on the impact of the number of ORB feature points on the computation time consumption.

the reproduced simulated affine transformations, thereby resulting in disparate final results. Consequently, the number of ORB feature points during the acquisition of the optimal simulation parameter group does not directly influence the precision of the method. Instead, it exerts an impact on the final precision by affecting the values of the optimal simulation parameter group and modifying the simulated transformations in the precise matching stage. Provided that the optimal simulated affine parameter group can be identified, consistent results will be obtained, as evidenced by the identical affine transformation results for cases such as 1-2, 1-3, 1-4, and 1-5.

Fig. 8 shows the impact of different numbers of ORB feature points on the running cost of our method. The experimental results indicate that the number of ORB feature points significantly affects the operation time. Considering its comprehensive impact on precision, too few points make the output simulation parameter group sub-optimal, reducing the method's precision. Meanwhile, too many points extend the running time without improving the precision further. Based on this, we suggest choosing 1,000 points for ORB feature extraction and matching when obtaining the optimal simulation parameter group.

To verify the effectiveness of adding the scale parameter in the simulated transformation, under the condition that 1,000 feature points were extracted by ORB, we directly calculated the matching accuracy and the number of matching point pairs in this process (both were the average values of the experimental results on five pairs of images), and counted the impact of different values of the scale parameter on the time consumption of the method proposed in this paper (the time consumption counted here was the average time consumption of the complete method on five pairs of images). The

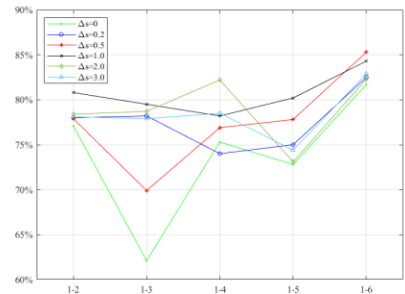

(a) The matching precision in the process of obtaining the optimal simulation parameter group.

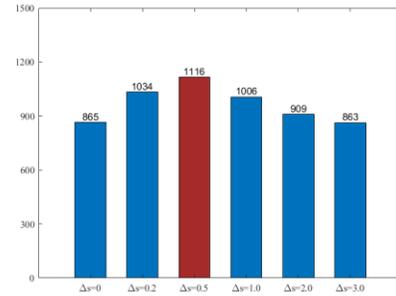

(b) The average number of matching point pairs in the process of obtaining the optimal simulation parameter group.

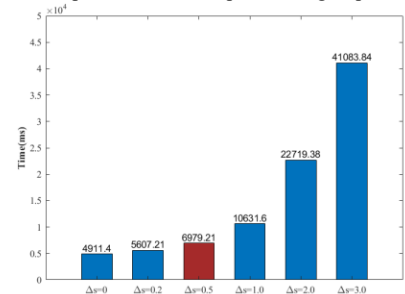

(c) The average time consumption of our method under different values of $\Delta s$.

**Fig. 9.** Investigation into the impact of $\Delta s$ values on the performance of the method in the Graffiti image set. In figures (b) and (c), the red bars represent the experimental results of our method with adopted parameters.

experimental results are shown in Fig. 9.

In Fig. 9(a), when $\Delta s = 0$, it represents the result without scale transformation. As observed, the experimental precision without scale change is mostly the worst. Also, a larger $\Delta s$ doesn't necessarily mean higher precision. Looking at the average number of matching point pairs in Fig. 9(b), we can see that under larger affine transformations, more point pairs are obtained when $\Delta s$ ranges from 0.2 to 1.0, leading to better experimental results. During the experiment, when $\Delta s$ is too large, the image gets bigger and more feature points are extracted. However, neither the precision nor the number of effective feature points keeps increasing. Instead, the computation time of the method increases significantly. Experimental tests show that when $\Delta s$ increases by 1.0 each time, the experimental time approximately doubles, as shown in Fig. 9(c). Incorporating scale change into our method doesn't directly change the final precision result but greatly affects the reliability of the optimal affine transformation parameter group obtained from the initial ORB matching. For example, considering the affine transformations of Image 1 to Image 5 in the Graffiti image set, when $\Delta s = 0$, the optimal longitude angle *bestϕ* = 178.191° in the optimal simulation parameter group, the optimal latitude angle *bestθ* = 70°, and the final precision is 88.99%. When Δ*s* = 0.5, *bestϕ* = 0°, *bestθ* = 70°, with a final precision of 91.15%. Consequently, when choosing the value of Δ*s*, we need to consider both its impact on precision and speed. Based on the experimental results in Fig. 9, our method performs better when Δ*s* = 0.5. Under large-angle affine transformations, the results are the most stable, and the impact on the computation time is relatively small. Therefore, Δ*s* = 0.5 is adopted in our experiments.

Combining the experimental results shown in Table 2, Fig. 8 and Fig. 9, and comprehensively considering the impact of the number of feature points extracted by ORB in the process of obtaining the optimal simulation parameter group as well as the value of the variation in the scale parameter on the precision and speed of our method, we recommend setting the number of ORB feature points to 1000 and the value of the variation in the scale parameter to Δ*s* = 0.5.

## V. Comparative Analysis of Methods

We compared the method proposed in this paper with algorithms such as SIFT, ORB, ASIFT, AORB, Qi-SIFT, and Fast-AASIFT in terms of running speed and matching precision.

### A. Complexity theory analysis

To demonstrate the superiority of our method in terms of computational speed, we analyzed the theoretical time consumption of our method as well as that of ASIFT and AORB. The process of the SIFT algorithm involves two rounds of image feature extraction and one brute-force matching process. The computational cost of SIFT feature extraction and matching is denoted as shown in (8).

$$C_{SIFT} = 2 \times O_1(m) + O_1(n^2) \quad (8)$$

Among them, $O_1(m)$ denotes the computational complexity of SIFT feature extraction which is related to the image size m, while $O_1(n^2)$ stands for the computational complexity of SIFT feature matching that is associated with the number of feature points n. In accordance with the sampling rules of ASIFT regarding longitude and latitude, the number of simulated images for a single image in ASIFT (with *n* = 5 adopted here) amounts to 45. Supposing that the size of the simulated images for both of the two images is *m*, and the average number of feature points is *n*, then the computational cost of the ASIFT algorithm is presented as shown in (9).

$$C_{ASIFT} = 45 \times 2 \times O_1(m) + 45^2 \times O_1(n^2) = 45 \times C_{SIFT} + 45 \times 44 \times O_1(n^2) \quad (9)$$

Apparently, considering that the computational time costs of SIFT's own feature extraction and matching are relatively high, the computational time of ASIFT is rather long and its immediacy is poor. For AORB, in the above computational cost, if the feature extraction and matching time of SIFT is replaced by the computational time cost of ORB, then the computational cost of AORB is shown in (10).

$$\begin{aligned} C_{AORB} &= 45 \times 2 \times O_2(m) + 45^2 \times O_2(n^2) \\ &= 45 \times C_{ORB} + 45 \times 44 \times O_2(n^2) \end{aligned} \quad (10)$$

In (10), $O_2(m)$, $O_2(n^2)$ respectively represent the computational complexities corresponding to ORB feature extraction and matching. Due to the fact that ORB utilizes the oFAST feature extraction method and the binary descriptor rBRIEF, the computational speed of ORB is much higher than that of SIFT. As a result, the computational cost of AORB is smaller than that of ASIFT.

According to the sampling rules and experimental settings in this paper, the actual computational cost of our method is shown in (11).

$$\begin{aligned} C_{Ours} &= 6 \times O_1(m) + 3 \times O_1(n^2) + 3 \times O_2(m) + 4 \times O_2\left((1+\Delta s)^2 m\right) \\ &+ 5 \times O_2\left((1+2\times\Delta s)^2 m\right) + 8 \times O_2\left((1+3\times\Delta s)^2 m\right) \\ &+ 10 \times O_2\left((1+4\times\Delta s)^2 m\right) + 15 \times O_2\left((1+5\times\Delta s)^2 m\right) + 45 \times O_2(n^2) \\ &\overset{\Delta s=0.5}{\simeq} 3 \times C_{SIFT} + 40 \times C_{ORB} + 5 \times O_2(n^2) \end{aligned} \quad (11)$$

It can be found according to (11) that the computational cost of our method has a higher correlation with the computational time of the ORB algorithm. Therefore, the number of ORB feature points suitable for our method is of relatively significant importance to the speed. Compared with ASIFT and AORB, our method can clearly demonstrate its superiority in terms of time. It greatly simplifies the steps and reduces the time for feature extraction and matching, has a lower computational cost, and can significantly improve the immediacy of the method. Moreover, the experimental results are basically consistent with our analysis.

### B. Comparative Analysis of Experimental Results

In this paper, comparative experiments were conducted on SIFT, ORB, ASIFT, AORB, Qi-SIFT, Fast-AASIFT and our method based on the Graffiti image set of the Mikolajczyk standard dataset, the Wall image set, the Adam painting image set of the MorelYu_2009 dataset, and the Magazine image set. Among them, the homography matrix of the Adam painting image set was derived from the v_adam image set of the H-Patches [38] dataset and the adam image pair in the EVD (Extreme View Dataset) [39] dataset. However, the Magazine image set lacks a homography matrix, so the proportion of inliers in RANSAC was used as the precision substitute instead. Firstly, there was the precision testing experiment, and the statistical situation of the experimental results of the accuracy rate is shown in Fig. 10.

As shown in Fig. 10, SIFT and ORB perform well when the viewpoint change is relatively small. Qi-SIFT has been improved based on SIFT and can achieve better results than SIFT. However, as the viewpoint change increases, its performance decays rapidly until it is almost zero. ASIFT and AORB have scale, rotation invariance and affine invariance, so their performance decay is relatively moderate. Nevertheless, when the viewpoint change is small, their accuracy performance is actually inferior to that of SIFT instead. Fast-AASIFT has an accuracy similar to that of ASIFT under large affine transformation angles, but its performance is rather poor when the viewpoint change is small. When the viewpoint change is small, the precision of our method is close to or even better than that of the SIFT method, and when the viewpoint change is large, it can also maintain a

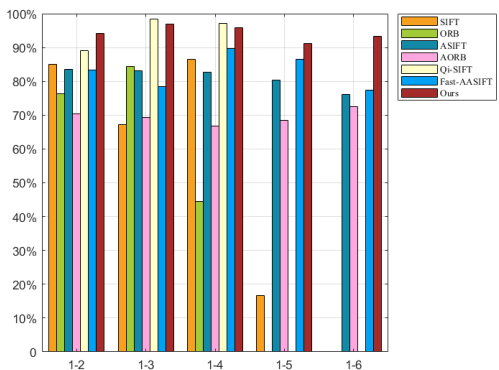

(a) The accuracy rates of various algorithms for the Graffiti image set.

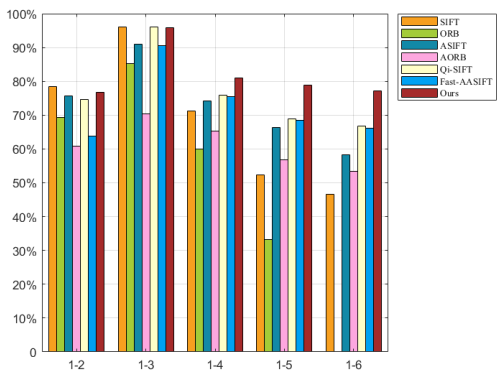

(b) The accuracy rates of various algorithms for the Wall image set.

**Fig. 10.** Comparative experiment on precision tests of different methods.

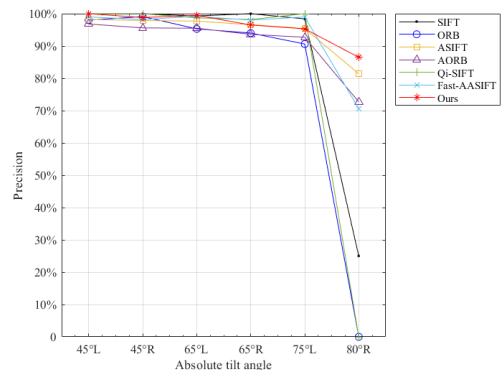

(a) Experimental results on the Adam painting image set ($Th = \sqrt{8}$).

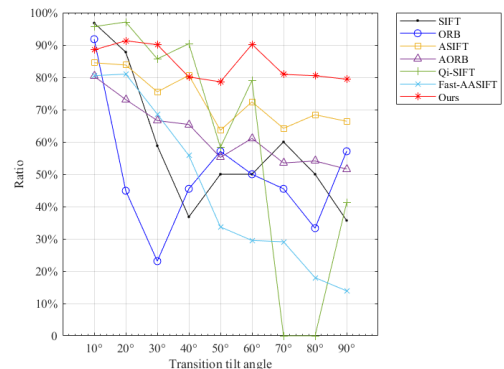

(b) Experimental results on the Magazine image set.

**Fig. 11.** The variation relationship between the precision of different algorithms and the tilt angle.

relatively high precision.

To further explore the relationship between the precision of various methods and the absolute tilt angle as well as the transition tilt angle, we conducted a supplementary precision experiment regarding the absolute tilt angle on the Adam painting image set and an experiment on the proportion of RANSAC inliers for the transition tilt angle on the Magazine image set. Among them, for all the methods running on the Magazine image set, the RANSAC step for finally screening feature points has been cancelled to avoid influencing the experimental results on this image set. The experimental results are shown in Fig. 11.

As shown in Fig. 11(a), when the absolute tilt angle is less than 75°, since the quality of image features is at a relatively high level, most methods can maintain a high precision. However, when the absolute tilt angle reaches 75°, the precision of most methods drops sharply. Among them, the precision of ORB and Qi-SIFT drops to 0 at 80°, and that of SIFT also decreases to about 25%. Only the precision of our method can always remain above 85%. Fig. 11(b) shows that when the transition tilt angle increases to 30°, the proportion of inliers of most methods will decline significantly. Affine invariant methods such as ASIFT, AORB, Fast-AASIFT and our method can still maintain a relatively high proportion of inliers, while the proportion of inliers of non-affine invariant methods such as SIFT, ORB and Qi-SIFT has already dropped to a relatively low range. Although the graph shows that the proportion of inliers of ORB and Qi-SIFT rises slightly in the range from 40° to 60°, in fact, in this range, the number of effective point pairs that can be extracted by the three non-affine invariant methods has already dropped to less than 10 pairs. At this time, the feature extraction methods have basically failed, and the proportion of inliers has no practical significance. When the angle is above 60°, except for our method, the proportion of inliers of other methods has dropped below 70%. Only our method can maintain a proportion of inliers of 80% or above from 10° to 80°.

Combining the results of Fig. 10 and Fig. 11, our method has the best performance among the compared methods, and its precision decays extremely slowly, demonstrating a stable and excellent performance.

Next, a comparative experiment on the computational speed was carried out between our method and affine invariant feature extraction methods such as ASIFT, AORB and Fast-AASIFT on the Graffiti image set and the Wall image set of the Mikolajczyk dataset. To meet the objectivity requirements of the experiment, the experimental results are the average time consumption of three groups of experimental test data, and the results are shown in Fig. 12.

As depicted in Fig. 12, the average time consumption of the method proposed in this paper is lower than that of Fast-AASIFT and is significantly less than that of ASIFT and AORB. Specifically, the computational speed of our method is 4 to 6 times faster than that of ASIFT and 7 to 10 times faster than that of AORB. Moreover, for images with more abundant features, the superiority of the proposed method in terms of speed becomes even more pronounced. The primary reasons

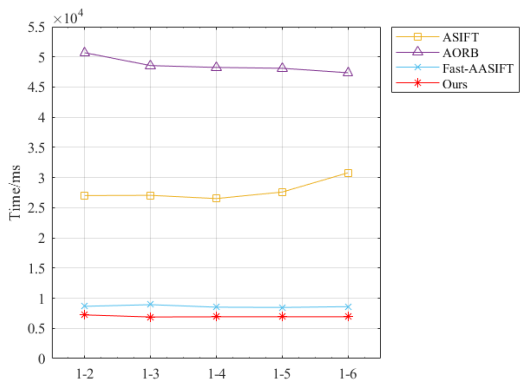
(a) Time consumption of different methods on the Graffiti image set.

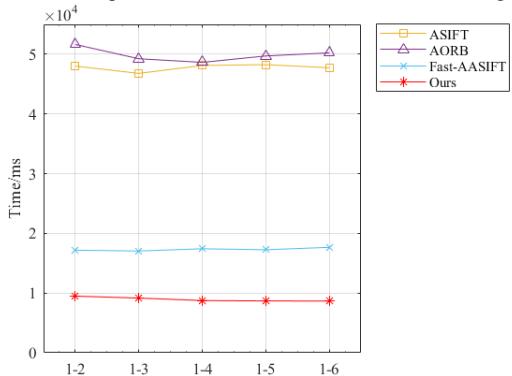
(b) Time consumption of different methods on the Wall image set.

**Fig. 12.** Experimental comparison of the computational time consumption of different algorithms.

TABLE III
STATISTICS TABLE OF MEMORY OCCUPANCY ON THE GRAFFITI IMAGE SET (UNIT: KB).

| Image pair | Memory(*Ours*)↓ | Memory(*ASIFT*)↓ | Percent(*Ours/ASIFT*)↓ |
|---|---|---|---|
| 1-2 | 244460 | 390480 | 62.60% |
| 1-3 | 247568 | 477828 | 51.81% |
| 1-4 | 252368 | 523164 | 48.24% |
| 1-5 | 246620 | 397656 | 62.02% |
| 1-6 | 249440 | 450916 | 55.32% |
| Average | 248091 | 448009 | 56.00% |

for this are as follows. Firstly, our method reduces the quantity of simulated images, thereby markedly decreasing the frequencies of feature extraction and matching. Secondly, in the initial matching phase, it employs ORB for feature extraction and matching and confines the number of feature points within a specific range, which contributes to a substantial reduction in the system's running time and operational costs. Theoretically, AORB utilizes the ORB feature extraction method and binary descriptors, and thus its computational time is expected to be shorter than that of ASIFT. However, experimental results reveal that OpenCV has implemented an acceleration approach for SIFT feature extraction and matching, which enables SIFT to operate at a faster pace. Consequently, when the number of feature points is approximately the same, ASIFT and AORB actually consume a comparable amount of time.

In addition, we also verified the memory usage of our method and ASIFT. We counted the running memory usage of five pairs of images on the Graffiti image set, as shown in Table 3. In actual operation, ASIFT occupies an average memory of approximately 448,009 KB, while our method occupies an average memory of approximately 248,091 KB, which is only 56% of that of ASIFT on average.

## VI. CONCLUSION

This paper proposes a feature point extraction method for affine images with large viewing angles, which takes into account both the improvement of precision and speed. This method has good affine invariance, and it remains effective as the absolute tilt angle of a single image increases (from 0 to the maximum of 80°) and as the transition angle between two images increases (from 0 to 90°). The experimental results show that the image matching precision of our method is 15% higher than that of the classic ASIFT. Meanwhile, its processing speed is more than 5 times that of ASIFT, and its memory occupancy is approximately 56% of that of ASIFT. Compared with the latest Fast-AASIFT, the average precision of the proposed method in this paper is about 10% higher, and its average time consumption is only 61.4% of that of Fast-AASIFT. The precision of feature extraction by our method decays extremely slowly with the increase of the angle, demonstrating a strong adaptability to extremely large-angle affine transformations.

REFERENCES

[1] H. P. Zhang, C. C. Leng, X. Yan, G. R. Cai, Z. Pei, N. G. Yu, and A. Basu, "Remote Sensing Image Registration Based on Local Affine Constraint With Circle Descriptor," *IEEE Geoscience and Remote Sensing Letters,* vol. 19, 2022.
[2] Y. Wu, W. P. Ma, Q. X. Su, S. D. Liu, and Y. H. Ge, "Remote sensing image registration based on local structural information and global constraint," *Journal of Applied Remote Sensing,* vol. 13, no. 1, FEB 19, 2019.
[3] J. Yan, X.-C. Yin, W. Lin, C. Deng, H. Zha, and X. Yang, "A Short Survey of Recent Advances in Graph Matching," in Proceedings of the 2016 ACM on International Conference on Multimedia Retrieval, New York, New York, USA, 2016, pp. 167–174.
[4] Y. Wu, W. P. Ma, M. G. Gong, L. Z. Su, and L. C. Jiao, "A Novel Point-Matching Algorithm Based on Fast Sample Consensus for Image Registration," *IEEE Geoscience and Remote Sensing Letters,* vol. 12, no. 1, pp. 43-47, JAN, 2015.
[5] W. P. Ma, Z. L. Wen, Y. Wu, L. C. Jiao, M. G. Gong, Y. F. Zheng, and L. Liu, "Remote Sensing Image Registration With Modified SIFT and Enhanced Feature Matching," *IEEE Geoscience and Remote Sensing Letters,* vol. 14, no. 1, pp. 3-7, JAN, 2017.
[6] D. Giveki, M. A. Soltanshahi, and M. Yousefvand, "Proposing a new feature descriptor for moving object detection," *Optik,* vol. 209, May, 2020.
[7] J. Gao, and Z. Sun, "An Improved ASIFT Image Feature Matching Algorithm Based on POS Information," *Sensors (Basel),* vol. 22, no. 20, Oct 12, 2022.
[8] B. Fan, Q. Kong, X. Wang, Z. Wang, S. Xiang, C. Pan, and P. Fua, "A performance evaluation of local features for image-based 3D reconstruction," *IEEE Transactions on Image Processing,* vol. 28, no. 10, pp. 4774-4789, 2019.
[9] J. Y. Ma, X. Y. Jiang, A. X. Fan, J. J. Jiang, and J. C. Yan, "Image Matching from Handcrafted to Deep Features: A Survey," *International Journal of Computer Vision,* vol. 129, no. 1, JAN, 2021.
[10] G. R. Cai, P. M. Jodoin, S. Z. Li, Y. D. Wu, S. Z. Su, and Z. K. Huang, "Perspective-SIFT: An efficient tool for low-altitude remote sensing image registration," *Signal Processing,* vol. 93, no. 11, pp. 3088-3110, Nov, 2013.

[11] K. Mikolajczyk, and C. Schmid, "Scale & Affine Invariant Interest Point Detectors," *International Journal of Computer Vision,* vol. 60, no. 1, pp. 63-86, Oct, 2004.
[12] J. Matas, O. Chum, M. Urban, and T. Pajdla, "Robust wide-baseline stereo from maximally stable extremal regions," *Image and Vision Computing,* vol. 22, no. 10, pp. 761-767, Sep 1, 2004.
[13] K. Mikolajczyk, T. Tuytelaars, C. Schmid, A. Zisserman, J. Matas, F. Schaffalitzky, T. Kadir, and L. van Gool, "A Comparison of Affine Region Detectors," *International Journal of Computer Vision,* vol. 65, no. 1-2, pp. 43-72, Nov, 2005.
[14] D. G. Lowe, "Distinctive image features from scale-invariant keypoints," *International Journal of Computer Vision,* vol. 60, no. 2, pp. 91-110, NOV 2004, doi: 10.1023/B:VISI.0000029664.99615.94.
[15] N. Tai, D. Wu, and J. Qi, "A Method to Extract High Robust Keypoints Based on Improved SIFT," *Acta Aeronautica et Astronautica Sinica,* vol. 33, no. 12, pp. 2313-2321, 2012.
[16] W. Peng, W. Ping, S. Zhenkang, G. A. O. Yinghui, and Q. U. Zhiguo, "A Novel Algorithm for Affine Invariant Feature Extraction Based on SIFT," *Signal Processing*, vol. 27, no. 1, pp. 88-93, 2011.
[17] J.-M. Morel, and G. Yu, "ASIFT: A New Framework for Fully Affine Invariant Image Comparison," *SIAM Journal on Imaging Sciences,* vol. 2, no. 2, pp. 438-469, 2009.
[18] Y. Hou, S. Zhou, L. Lei, and J. Zhao, "Fast fully affine invariant image matching based on ORB," *Computer Engineering and Science,* vol. 36, no. 2, pp. 303-310, 2014.
[19] K. Su, G. Han, and H. Sun, "Anti-Viewpoint Changing Image Matching Algorithm Based on SURF," *Chinese Journal of Liquid Crystals and Displays,* vol. 28, no. 4, pp. 626-632, 2013.
[20] Y. W. Pang, W. Li, Y. Yuan, and J. Pan, "Fully affine invariant SURF for image matching," *Neurocomputing,* vol. 85, pp. 6-10, May 15, 2012.
[21] C. Fu, L. Deng, G. Lu, and M. Fei, "Improved image matching based on fast retina keypoint algorithm," *Computer Engineering and Application,* vol. 52, no. 19, pp. 208-212, 2016, 2016.
[22] X. M. Ma, Y. Yang, Y. M. Yi, L. Zhu, and M. Dong, "A computationally efficient affine-invariant feature for image matching with wide viewing angles," *Optik,* vol. 247, Dec, 2021.
[23] E. Rublee, V. Rabaud, K. Konolige, and G. Bradski, "ORB: An efficient alternative to SIFT or SURF," in *2011 International Conference on Computer Vision*, 6-13 Nov. 2011 2011, pp. 2564-2571, doi: 10.1109/ICCV.2011.6126544.
[24] H. Bay, T. Tuytelaars, and L. Van Gool, "SURF: Speeded up robust features," *Computer Vision - ECCV 2006, Pt 1, Proceedings,* vol. 3951, pp. 404-417, 2006.
[25] H. Bay, A. Ess, T. Tuytelaars, and L. Van Gool, "Speeded-Up Robust Features (SURF)," *Computer Vision and Image Understanding,* vol. 110, no. 3, pp. 346-359, Jun, 2008.
[26] A. Alahi, R. Ortiz, and P. Vandergheynst, "Freak: Fast retina keypoint," in *2012 IEEE conference on computer vision and pattern recognition*, 2012: Ieee, pp. 510-517
[27] K. Turkowski, "Filters for common resampling tasks," *Graphics gems*, pp. 147–165: Academic Press Professional, Inc., 1990.
[28] K. M. Flegal, B. K. Kit, H. Orpana, and B. I. Graubard, "Association of All-Cause Mortality With Overweight and Obesity Using Standard Body Mass Index Categories A Systematic Review and Meta-analysis," *JAMA-Journal of the American Medical Association,* vol. 309, no. 1, pp. 71-82, JAN 2, 2013.
[29] K. Mikolajczyk, T. Tuytelaars, C. Schmid, A. Zisserman, J. Matas, F. Schaffalitzky, T. Kadir, and L. van Gool, "A Comparison of Affine Region Detectors," *International Journal of Computer Vision,* vol. 65, no. 1-2, pp. 43-72, Nov, 2005.
[30] X. M. Ma, D. Liu, J. Zhang, and J. Xin, "A fast affine-invariant features for image stitching under large viewpoint changes," *Neurocomputing,* vol. 151, pp. 1430-1438, Mar 3, 2015.
[31] L. Yu, Q. Yi, M. Jin, and K. Zhou, "Geometry and Bionic Fusion Feature Extraction Method for Affine Target Recognition," *Acta Electronica Sinica,* vol. 51, no. 6, pp. 1607-1618, 2023.
[32] F. Z. Zhu, J. C. Li, B. Zhu, H. L. Li, and G. X. Liu, "UAV remote sensing image stitching via improved VGG16 Siamese feature extraction network," *Expert Systems with Applications,* vol. 229, NOV 1, 2023.
[33] J. Yue, S.-l. Gao, F.-m. Li, and N.-b. Cai, "Fast image matching algorithm with approximate affine and scale invariance," *Optics and Precision Engineering,* vol. 28, no. 10, pp. 2349-2359, 2020.
[34] J. Gao, and Z. Sun, "An Improved ASIFT Image Feature Matching Algorithm Based on POS Information," *Sensors (Basel),* vol. 22, no. 20, Oct 12, 2022.
[35] F. Qi, Y. Wan, and Y. Li, "An Improved SIFT Algorithm for Large Tilt Angle Images," in 2020 IEEE 9th Joint International Information Technology and Artificial Intelligence Conference (ITAIC), 2020, pp. 958-962.
[36] S. Leutenegger, M. Chli, and R. Y. Siegwart, "BRISK: Binary Robust invariant scalable keypoints," in 2011 *International Conference on Computer Vision*, 2011, pp. 2548-2555.
[37] W. Ji, and F. Yang, "Affine medical image registration with fusion feature mapping in local and global," *Physics in Medicine and Biology,* vol. 69, no. 5, MAR 7, 2024.
[38] V. Balntas, K. Lenc, A. Vedaldi, T. Tuytelaars, J. Matas, and K. Mikolajczyk, "H-Patches: A Benchmark and Evaluation of Handcrafted and Learned Local Descriptors," *IEEE Conference on Computer Vision and Pattern Recognition,* vol. 42, no. 11, pp. 2825-2841, NOV 2020, doi: 10.1109/TPAMI.2019.2915233.
[39] D. Mishkin, J. Matas, and M. Perdoch, "MODS: Fast and robust method for two-view matching," *Computer Vision and Image Understanding,* vol. 141, pp. 81-93, 2015.
[40] M. Zhang, H. Jin, Z. Xiao, and C. Guillemot, "A Light Field FDL-HCGH Feature in Scale-Disparity Space," *IEEE Transactions on Image Processing,* vol. 31, pp. 6164-6174, 2022.
[41] M. Moyou, A. Rangarajan, J. Corring, and A. M. Peter, "A Grassmannian Graph Approach to Affine Invariant Feature Matching," *IEEE Transactions on Image Processing,* vol. 29, pp. 3374-3387, 2020.
[42] C. Cui, and K. N. Ngan, "Scale- and Affine-Invariant Fan Feature," *IEEE Transactions on Image Processing,* vol. 20, no. 6, pp. 1627-1640, 2011.
[43] Xinrong Mao, Kaiming Liu, and Yanfen Hang. 2020. Feature Extraction and Matching of Slam Image Based on Improved SIFT Algorithm. In Proceedings of the 2020 2nd Symposium on Signal Processing Systems (SSPS '20). Association for Computing Machinery, New York, NY, USA, 18–23. https://doi.org/10.1145/3421515.3421528.
[44] Wengang Zhou, Houqiang Li, Yijuan Lu, and Qi Tian. 2013. SIFT match verification by geometric coding for large-scale partial-duplicate web image search. ACM Trans. Multimedia Comput. Commun. Appl. 9, 1, Article 4 (February 2013), 18 pages. https://doi.org/10.1145/2422956.2422960.
[45] Qiaohui Liu. 2022. Smooth Stitching Method of VR Panoramic Image Based on Improved SIFT Algorithm. In Proceedings of the Asia Conference on Electrical, Power and Computer Engineering (EPCE '22). Association for Computing Machinery, New York, NY, USA, Article 49, 1–6. https://doi.org/10.1145/3529299.3531499.
[46] Changyong Guo, Zhaoxin Zhang, Jinjiang Li, Xuesong Jiang, Jun Zhang, and Lei Zhang. 2020. Robust Visual Tracking Using Kernel Sparse Coding on Multiple Covariance Descriptors. ACM Trans. Multimedia Comput. Commun. Appl. 16, 1s, Article 20 (January 2020), 22 pages. https://doi.org/10.1145/3360308